%% file: main.tex
\documentclass{edm_article}

\usepackage[T1]{fontenc}
\usepackage{textcomp}

\usepackage{booktabs}
\usepackage{multicol}
\usepackage{multirow}
\usepackage{amsmath}
\usepackage{bbm}
\usepackage{verbments}

\usepackage[most,minted]{tcolorbox}
\usepackage{graphicx}
\usepackage{tikz}
\tcbuselibrary{skins}
\usepackage{xcolor}
\usepackage{listings}
\usepackage{url}

\usepackage{fontawesome5}
\usepackage{hyperref}

\renewcommand{\arraystretch}{1.1}

\begin{document}

\title{Teaching Language Models How to Code Like Learners: Conversational Serialization for Student Simulation}

\numberofauthors{3}
\author{
\alignauthor Charles Koutcheme\\
       \affaddr{Aalto University}\\
       \affaddr{Espoo, Finland}\\
       \email{\large charles.koutcheme@aalto.fi}
\alignauthor Juho Leinonen\\
       \affaddr{Aalto University}\\
       \affaddr{Espoo, Finland}\\
       \email{\large juho.2.leinonen@aalto.fi}
\alignauthor Arto Hellas\\
       \affaddr{Aalto University}\\
       \affaddr{Espoo, Finland}\\
       \email{\large arto.hellas@aalto.fi}
}  

\maketitle

\begin{abstract}
\input{sections/00_abstract}
\end{abstract}


\input{sections/01_introduction}

\input{sections/02_related_work}
\input{sections/03_methodology}
\input{sections/04_experiments}

\input{sections/05_results}
\input{sections/06_concluding_discussion}
\input{sections/07_acknowledgements}

\bibliographystyle{abbrv}
\bibliography{main}  

\balancecolumns

\end{document}

%% file: sections/00_abstract.tex
Artificial students—models that simulate how learners act and respond within educational systems—are a promising tool for evaluating tutoring strategies and feedback mechanisms at scale. However, most existing approaches rely on prompting large, proprietary language models, limiting adaptability to specific courses and raising concerns around privacy, cost, and dependence. In this work, we propose a framework for training open-weight artificial programming learners directly from authentic student process data. Our approach serializes temporal log traces into a conversational format, representing each student's problem-solving process as a dialogue between the learner and their automated assessment system. Student code submissions and environment feedback, such as test outcomes, grades, and error traces, form alternating conversational turns, enabling models to learn from the iterative debugging process. We additionally introduce a training pipeline combining supervised fine-tuning with preference optimization to align models with authentic student debugging behavior. We evaluate our framework by training Qwen models at 4B and 8B scales on a large-scale dataset of real student submissions to Python programming assignments. Our results show that incorporating environment feedback strengthens models' ability to replicate student debugging behavior, improving over both prior code-only approaches and prompted large language models baselines in functional alignment and code similarity. We release our code to support reproducibility.

%% file: sections/01_introduction.tex
\section{Introduction}

To better support learners at scale, computing education research has long relied on models of students built from the rich log data collected as they solve programming assignments~\cite{jadud2006methods,watson2013predicting}. Much of this work focuses on capturing what students know, such as estimating mastery over concepts through knowledge tracing~\cite{corbett1994knowledge,kasurinen2009estimating,wang2017deep}, or identifying misconceptions that explain incorrect outcomes~\cite{sirkia2012exploring}. 
These models provide valuable insights into learning progress and have informed the design of successful interventions~\cite{azcona2017targeting}. 
However, while such approaches are effective for analyzing student cognition and predicting outcomes, operationalizing interventions based on these models remains challenging. In many settings, what is needed is not only an estimate of what a student knows, but a model of how a student behaves: how they act, react, and progress when interacting with learning systems. This has motivated growing interest in \emph{artificial students}—models that act as stand-ins for real learners by generating plausible sequences of actions~\cite{vanlehn1994applications,matsuda2007evaluating,brown1999simulated}. When realistic, artificial students could enable the simulation of student–system interactions and support controlled experimentation, such as comparing tutoring strategies, evaluating feedback policies, or assessing the difficulty of programming tasks, as already done in prior works~\cite{phung2024automatinghumantutorstyleprogramming,dinucu-jianu-etal-2025-problem,PyTaskSyn}, without requiring repeated classroom deployments. 
Despite this promise, existing artificial student models remain limited, particularly in programming education. Recent work using large language models has shown that it is possible to generate student-like code~\cite{leinonen2025llmitation,macneil2024synthetic}, but these approaches often rely on prompting alone and tend to reproduce surface-level error patterns rather than the dynamic nature of real student debugging. In particular, existing methods remain limited in their ability to realistically simulate how learners tackle programming assignments.

In this work, we leverage the rich process data already collected by programming learning environments to train small and medium size open-weight models for \emph{within-assignment student simulation}~\cite{miroyan2025parastudentgeneratingevaluatingrealistic,ross2025modelingstudentlearning38}. Since student data is sensitive and institutional, this offers a more practical path than prompting proprietary systems~\cite{ross2025modelingstudentlearning38}, keeping data local while learning the iterative debugging patterns that prompting alone cannot reliably elicit.
Our core innovation is to represent each problem-solving trajectory as a dialogue between a student and their learning environment. Student code submissions are encoded as assistant turns, while outputs from automated assessment systems (e.g., unit-test outcomes, or grader messages) form the corresponding user turns. By adopting the conversational format commonly used to interact with language models, we turn raw log traces into structured sequences that capture how a student’s code evolves in response to feedback. This representation enables chat-capable language models to learn from iterative debugging sequences without requiring architectural modifications.

To train artificial students from this conversational representation, we propose a pipeline combining supervised fine-tuning with offline preference optimization via Direct Preference Optimization~\cite{rafailov2024direct}. We additionally explore online preference optimization methods based on Group Relative Preference Optimization~\cite{yu2025dapoopensourcellmreinforcement}.
We evaluate our framework on FalconCode~\cite{falconcode}, a large-scale dataset of student submissions to Python programming assignment, training Qwen models~\cite{yang2025qwen3technicalreport} at 4B and 8B scales.
Our results show that incorporating environment feedback strengthens models' ability to replicate student debugging behavior, improving over both prior code-only baselines and prompted large language models in functional alignment and code similarity.

\noindent\textbf{Summary contributions.} Our main contributions are:
\vspace{-0.2cm}
\begin{itemize}
    \item \textbf{Student-environment serialization.} We introduce a conversational formulation of student–environment interactions that enables language models to learn students iterative debugging processes.

    \item \textbf{A training methodology for artificial students.} We propose a pipeline combining supervised fine-tuning and preference optimization to align language models with authentic student coding trajectories.

    \item \textbf{Empirical evaluation on real student data.} We validate our framework on a large-scale dataset of student programming submissions, showing that environment feedback and preference optimization yield complementary gains in simulation realism. We also release our code: \faGithub $\,$ \href{https://github.com/KoutchemeCharles/edm-conv-ser}{\nolinkurl{KoutchemeCharles/edm-conv-ser}}
\end{itemize}

%% file: sections/02_related_work.tex
\section{Related Work}

A complementary line of work focuses on \emph{knowledge tracing} (KT), whose primary goal is to estimate a student's mastery of knowledge components across exercises. While early approaches focused exclusively on predicting success on future assignments, more recent methods leverage language-model--based architectures to predict students' first submission to following assignments and the associated outcomes~\cite{liu-etal-2022-open,duan2025testcase,duan2025automatedknowledgecomponentgeneration}. While such methods may predict student programs, this prediction serves mainly as an intermediate signal for updating knowledge estimates across programming tasks, not as a goal in itself. In contrast, we aim to obtain models that can be unrolled over multiple steps within a single assignment to generate and simulate problem-solving trajectories.
Closer to our work in the educational domain is Miroyan et al.~\cite{miroyan2025parastudentgeneratingevaluatingrealistic}, who train open-weight language models using supervised fine-tuning to predict a student's next submission to Python programming assignments given a limited context of prior attempts. Their work shows that trained language models perform better than prompted LLMs. Ross et al.~\cite{ross2025modelingstudentlearning38} go further by fine-tuning pre-trained language models on 3.8 million traces from a block-based educational programming platform, showing that models trained on real edit sequences yield richer representations of student behavior than those trained on final programs alone. However, neither approach incorporates the learning environment's responses to student submissions.
Yet, serializing structured data into conversational format for language model training is well-established in NLP, with extensive work on synthetic dialogue generation~\cite{mukherjee2023orca,ding2023enhancing,xu2024wizardlm} and math-informed dialogues~\cite{akter2025mind}.

Closest to our approach is OpenCodeInterpreter~\cite{opencodeinterpreter}, which fine-tunes open-weight language models on multi-turn code-execution dialogues to improve code generation and refinement abilities. While their approach also serializes execution feedback into multi-turn dialogs, our work differs in two key aspects. First, our goal is to reproduce student problem-solving behavior, not to improve code generation; we train on authentic student trajectories rather than synthetic expert-generated interactions refined through execution outputs. Second, we develop preference optimization methods specifically adapted to student programming trajectories, leveraging infrastructure tied to learning environments, in particular the autograder, to construct preference pairs for training.

%% file: sections/03_methodology.tex
\section{From Logs to Dialogs}

In this section, we introduce our approach for transforming student log data into suitable data for student simulation. Our work assumes that students interact with an automated assessment system returning summative feedback~\cite{Butler1995FeedbackAS}.

\subsection{Assumptions}
 
We assume access to a deterministic grading function which, for a submitted program $a$ to a programming assignment \( d \), can return two observable components: \(v^{(d)}_r(a) = r\) and \(v^{(d)}_f(a) = o\). The numerical score \( r \in [0,1]\) measures the correctness of the executed program (e.g., proportion of passed test cases). The textual feedback \( o \) contains the textual outcome of executing that submission: summaries of passed tests, numerical grades, and/or even runtime error traces, depending on what each dataset provides. We do not assume any internal structure of the (summative) feedback, only that it is consistently produced and directly observable in the logs.
We also assume access to a dataset of institutional log data containing sequences of graded submissions and their outcomes: 
\(
\mathcal{D} = \{ (d^i, s^i) \}_{i=1}^N,
\)
where \( d^i \) denotes an assignment textual description and
\(
s^i = (u^i_1, u^i_2, \ldots, u^i_{T_i})
\)
denotes the sequence of graded interactions for a single student on that assignment. Each entry
\(
u^i_t = (a^i_t, o^i_t)
\)
contains the submitted program \( a^i_t \) and its outcome
\(
o^i_t = v^{(d^i)}_f(a^i_t).
\)
The trajectory length \( T_i \) equals the number of graded submissions recorded for assignment \( d^i \). 

\subsection{Serialization}
We represent student programming logs using the standard conversational format used to train and interact with chat-capable language models~\cite{ouyang2022training}.
We structure each trajectory as a dialogue where the \texttt{user} role represents the automated assessment system and the \texttt{assistant} role represents a student learning how to program. The \texttt{system} prompt establishes the context: the model is a novice student learning to program, solving assignments while interacting with a learning environment that provides summative feedback. The initial \texttt{user} message provides the assignment description $d_i$ and any relevant context. 
Each logged interaction $u^i_{t} = (a^i_t, o^i_t)$ then maps to two conversational turns: the student's code submission $a^i_t$ appears as an \texttt{assistant} turn, followed by the grading environment's feedback $o^i_t$ as the next \texttt{user} turn. This alternating structure naturally captures the iterative debugging process: students write code, receive feedback, then revise based on that feedback.
Below is an illustration using a simple Python assignment:

\vspace{0.2cm}
\input{prompt/dialog}

\vspace{0.2cm}

\section{Training Artificial Learners}

In this section, we present our training pipeline for training a language model $\pi_{\theta}$ to simulate how programming students solve assignments. Our core pipeline combines supervised fine-tuning with offline preference optimization on a serialized dataset $\mathcal{D}$. We additionally explore online preference optimization as an alternative to the offline stage.

\subsection{Supervised Fine-tuning}
 
The first method is to simply supervised finetune our model $\pi_{\theta}$ using the negative log likelihood (i.e., SFT) objective:
\[
\mathcal{L}_{\mathrm{SFT}}(\theta, \mathcal{D})
=
-\sum_{s^i \in \mathcal{D}}
\sum_{t=1}^{T_i}
\log \pi_{\theta}\!\left(
a^i_t \mid u^i_{<t}
\right).
\]
Following prior work, we only backpropagate the loss on assistant turns ~\cite{dettmers2023qloraefficientfinetuningquantized} containing student code submissions. The supervised fine-tuning step enables models to quickly learn patterns of student problem-solving.

\subsection{Offline Preference Optimization}

We use Direct Preference Optimization~\cite{rafailov2024direct} (DPO), an offline preference optimization algorithm that aligns language models using pairwise preference datasets. This algorithm has shown much success in applied AI in education~\cite{koutcheme2026aligning,woodrow2025dpo,scarlatos2024improvingvalidityautomaticallygenerated}.  
In this work, we construct preference datasets by forming contrastive pairs of candidate continuations from the same student trajectory. Let $u^i_{\le t}$ denote a sampled partial trajectory of a student up to step $t$. The preferred continuation is the student's immediate next submission $a^i_{t+1}$, while the dispreferred continuation is the first following submission with a different grade. Formally, if we define:
\[
k^{\star}(i,t)
=
\min\Big\{
k \in [2, T_i - t]:
v^{(d)}_r(a^i_{t+k}) \neq v^{(d)}_r(a^i_{t+1})
\Big\}.
\]
then the resulting preference dataset is :
\[
\mathcal{D}_{\text{DPO}}
=
\Big\{
\big(
u^i_{\le t},
a^i_{t+1},
a^i_{t+k^{\star}(i,t)}
\big)
\Big\}_{\substack{
i=1,\dots,N; \;
t\in\mathcal{T}_i
}}.
\]
We believe $a^i_{t+k^{\star}(i,t)}$, the nearest semantically distinct code the student actually wrote, provides a high-quality contrast that teaches the model \emph{why} students submit specific solutions at each step, preventing models from jumping prematurely to fully correct solutions, a pattern observed in prior work~\cite{miroyan2025parastudentgeneratingevaluatingrealistic}.
Using the resulting dataset, we optimize our model using the DPO loss~\cite{rafailov2024direct}:
\begin{align}
\label{eq:dpo_pref}
\mathcal{L}_{\mathrm{DPO}}(\theta)
&=
- \mathbb{E}_{(u^i_{\le t}, a^i_{t+1}, a^i_{t+k}) \sim \mathcal{D}_{\text{DPO}}} \nonumber\\
&\quad
\Big[
    \log \sigma\!\big(
    r_\theta(u^i_{\le t}, a^i_{t+1})
    - r_\theta(u^i_{\le t}, a^i_{t+k^{\star}(i,t)})
    \big)
\Big]
\end{align}
where $\sigma$ is the sigmoid function and
\begin{equation}
\label{eq:implicit_reward}
r_\theta(u,a)
\;\triangleq\;
\beta \log \frac{\pi_\theta(a \mid u)}{\pi_{\text{ref}}(a \mid u)}.
\end{equation}

with $\pi_{\theta}$ being the model being optimized and $\pi_{\text{ref}}$ being the reference policy (the model before the start of the optimization (often $\pi_{\text{SFT}}$)). In essence, Equation~\eqref{eq:dpo_pref} penalizes the model based on how much more it prefers the dispreferred generation over the preferred one using an implicit reward (Equation~\eqref{eq:implicit_reward}), defined by the model probabilities, with $\beta$ controlling how much the trained model weights $\theta$ can deviate from their initialization point (e.g., the SFT weights).

\subsection{Online Preference Optimization}

We additionally explore whether online preference optimization can exceed offline methods for student simulation. Specifically, we use a variant of Group Relative Preference Optimization (GRPO)~\cite{Guo2025}, an online algorithm that iteratively (1)~samples multiple candidate next-step programs from the current policy, (2)~evaluates them through a reward function, and (3)~updates the model according to their relative quality.
At each iteration, given a trajectory prefix \(u^i_{\le t}\), we sample \(G=4\) candidate next-step programs using top\_p sampling. Each candidate is scored using a discrete reward that captures two complementary aspects~\cite{koutcheme-etal-2025-direct} of next-step program prediction: functional alignment and syntactic similarity. Candidates whose abstract syntax tree exactly matches the student's actual next submission receive the highest reward (+2.0). Candidates that do not match exactly but achieve the same grade as the student's next attempt receive a smaller reward (+1.0). Candidates that fail to compile are penalized (-1.0), while all remaining candidates receive no reward. This tiered design prioritizes reproducing the student's specific solution while still rewarding functionally equivalent behavior, and aligns with outcome-based reward practices in code generation~\cite{le_coderl_2022}. We update model parameters using DAPO~\cite{yu2025dapoopensourcellmreinforcement}, a more stable version of the original GRPO algorithm. Due to space constraints, we refer the reader to the original paper and our code base for the full loss formulation.

%% file: prompt/dialog.tex
\tcbset{inputbox/.style={
  enhanced,
  colback=gray!3,
  colframe=gray!30!black,
  boxrule=0.18mm,
  left=2mm, right=2mm, top=0.5mm, bottom=0.5mm,
  arc=0.9mm,
  fontupper=\scriptsize,
  before skip=1mm,
  after skip=0.5mm,
}}

\newtcolorbox{systemmsg}[1][]{
  enhanced,
  colback=violet!5!white, colframe=violet!35!black,
  boxrule=0.18mm, left=2mm, right=2mm, top=0.5mm, bottom=0.3mm,
  arc=0.9mm, fontupper=\scriptsize,
  before skip=1mm, after skip=0.5mm,
  overlay={
    \node[anchor=north east, fill=violet!35!black, text=white,
          font=\bfseries\scriptsize, rounded corners=0.5mm,
          inner xsep=2mm, inner ysep=0.5mm]
          at ([xshift=0mm,yshift=0.7mm]frame.north east)
          {System};
  }
}

\newtcolorbox{usermsg}{
  enhanced,
  colback=white!5!white, colframe=blue!35!black,
  boxrule=0.18mm, left=2mm, right=2mm, top=0.5mm, bottom=0.5mm,
  arc=0.9mm, fontupper=\scriptsize,
  before skip=1mm, after skip=1mm,
  overlay={
    \node[anchor=north east, fill=blue!35!black, text=white,
          font=\bfseries\scriptsize, rounded corners=0.5mm,
          inner xsep=2mm, inner ysep=0.5mm]
          at ([xshift=0mm,yshift=0.7mm]frame.north east)
          {User};
  }
}

\newtcolorbox{assistantmsg}{
  enhanced,
  colback=gray!5!white, colframe=gray!40!black,
  boxrule=0.18mm, left=2mm, right=2mm, top=0.5mm, bottom=0.5mm,
  arc=0.9mm, fontupper=\scriptsize,
  before skip=1mm, after skip=1mm,
  overlay={
    \node[anchor=north east, fill=gray!40!black, text=white,
          font=\bfseries\scriptsize, rounded corners=0.5mm,
          inner xsep=2mm, inner ysep=0.5mm]
          at ([xshift=0mm,yshift=0.7mm]frame.north east)
          {Assistant};
  }
}

\definecolor{pykeyword}{RGB}{0, 0, 180}      
\definecolor{pystring}{RGB}{0, 130, 60}      
\definecolor{pycomment}{RGB}{120, 120, 120}  
\definecolor{pybuiltin}{RGB}{140, 60, 160}   
\definecolor{pynumber}{RGB}{180, 90, 0}      

\lstdefinestyle{dialog}{
  basicstyle=\ttfamily\scriptsize,
  breaklines=true,
  breakatwhitespace=false,
  columns=fullflexible,
  keepspaces=true,
  showstringspaces=false,
  frame=none,
  aboveskip=0pt,
  belowskip=0pt,
  xleftmargin=0pt,
  framexleftmargin=0pt,
}

\lstdefinestyle{dialogpy}{
  style=dialog,
  language=Python,
  keywordstyle=\color{pykeyword}\bfseries,
  stringstyle=\color{pystring},
  commentstyle=\color{pycomment}\itshape,
  emph={self,True,False,None},
  emphstyle=\color{pybuiltin}\bfseries,
  literate=%
    {0}{{{\color{pynumber}0}}}1
    {1}{{{\color{pynumber}1}}}1
    {2}{{{\color{pynumber}2}}}1
    {3}{{{\color{pynumber}3}}}1
    {4}{{{\color{pynumber}4}}}1
    {5}{{{\color{pynumber}5}}}1
    {6}{{{\color{pynumber}6}}}1
    {7}{{{\color{pynumber}7}}}1
    {8}{{{\color{pynumber}8}}}1
    {9}{{{\color{pynumber}9}}}1,
}

\lstset{style=dialog}

\begin{systemmsg}
\begin{lstlisting}
You are a first-year novice student learning 
programming in Python. Solve the given programming 
assignment(s). You will be interacting with a learning 
environment which will provide you with summative feedback.
\end{lstlisting}
\end{systemmsg}

\begin{usermsg}
\begin{lstlisting}
Write a Python function called "compute_average". 
The function should take as argument a (non-empty) list of 
integers and returns the mean over those elements. 
\end{lstlisting}
\end{usermsg}

\begin{assistantmsg}
\begin{lstlisting}[style=dialogpy]
def compute_average(nums):
    total = nums[0]
    for i in nums:
        total += i
    return average / len(nums)
\end{lstlisting}
\end{assistantmsg}

\begin{usermsg}
\begin{lstlisting}
Runtime error: undefined variable "average".
\end{lstlisting}
\end{usermsg}

\begin{assistantmsg}
\begin{lstlisting}[style=dialogpy]
def compute_average(nums):
    total = nums[0]
    for i in nums:
        total += i
    return total / len(nums)
\end{lstlisting}
\end{assistantmsg}

\begin{usermsg}
\begin{lstlisting}
Tests passed: 1/8
\end{lstlisting}
\end{usermsg}

\begin{assistantmsg}
\begin{lstlisting}[style=dialogpy]
def compute_average(nums):
    total = 0
    for i in nums:
        total += i
    return total / len(nums)
\end{lstlisting}
\end{assistantmsg}

\begin{usermsg}
\begin{lstlisting}
Tests passed: 8/8
\end{lstlisting}
\end{usermsg}

%% file: sections/04_experiments.tex
\section{Experiments}

In this section, we detail the components of our experiments aimed at evaluating the utility of our framework.

\subsection{Dataset}

We evaluate our framework using FalconCode~\cite{falconcode}, a large-scale CS1 Python programming dataset from the United States Air Force Academy.
The dataset includes student submissions, the associated grades, and the course autograder, which we used to regenerate the same detailed summative feedback students received during the course, showing pass/fail and expected outputs per test case.
FalconCode contains three assignment difficulty levels: ``skill'', ``lab'', and ``project''. We excluded project assignments (lacking automated tests) and skill assignments (1–10 lines of code). Lab assignments represent medium-sized programs (10–50 lines). To manage the costs of running our experiments, we arbitrarily selected the assignment with the highest number of submissions from each lesson regrouping assignments targetting similar concepts.
We retained trajectories where students passed at least one test, as all-failing trajectories often reflect non-serious attempts. We further removed non-compiling submissions but retained runtime errors and traces as environment feedback. We leave the handling of syntax errors to future work.
FalconCode spans three semesters. To mimic realistic deployment, we use data from Spring 2021 for training, and evaluate on a randomly sampled susbet of 1000 trajectories from Spring 2022.

\subsection{Models}

We fine-tune two open-weight language models from the Qwen family~\cite{yang2025qwen3technicalreport}: \texttt{Qwen3-4B-Instruct} and \texttt{Qwen3-8B}. We selected these models for their strong performance on coding benchmarks and to study model size effects on student simulation quality.

\textbf{Model variants.}
We evaluate variants to isolate the contribution of our framework components. 
As a baseline to prior work~\cite{miroyan2025parastudentgeneratingevaluatingrealistic}, we train models \emph{without} environment feedback (\textbf{PARA}).
Within our framework, we compare \textbf{SFT} (supervised fine-tuning), \textbf{DPO} (offline preference optimization), and \textbf{DAPO} (online preference optimization). For both preference optimization, we start training from the SFT weights. 

\textbf{Training setup.}
We use \texttt{unsloth}~\cite{unsloth} to enable long-context fine-tuning on NVIDIA V100 GPUs with 32\,GB of VRAM, using \texttt{triton}, our institution research cluster.
All models are fine-tuned with QLoRA~\cite{dettmers2023qloraefficientfinetuningquantized}, following established hyperparameter recommendations~\cite{schulman2025lora,unsloth_lora_hyperparameters}: rank $r=8$, $\alpha=16$.
Models are trained with a maximum context length of 4096 tokens. When trajectories exceed this limit, we remove earlier assistant turns until the sequence fits, creating a dynamic sliding window over recent submissions. 
For PARA models, unlike~\cite{miroyan2025parastudentgeneratingevaluatingrealistic} who use a fixed window of $k=4$ prior submissions, we dynamically limit prior submissions based on the context window size such that all trained models operate under the same computational constraints.
We use a cosine learning rate of $1\times10^{-4}$ for SFT variants (including PARA) 
(one epoch) and $1\times10^{-5}$ for preference optimization, 
with $\beta=0.5$ for DPO following~\cite{scarlatos2024improvingvalidityautomaticallygenerated}.
For SFT variants and DPO, we select the best checkpoint based on validation loss from 20\% of training data.
For DPO, we sample all positions from all available trajectory prefixes from the training set. For GRPO, we sample two positions from each trajectory due to computational costs, generating $G=4$ possible attempts per trajectory.

\textbf{In-context learning baselines.}
To assess the value of fine-tuning, we include in-context learning (ICL) baselines using \textbf{BASE} model variants without training. We also evaluate the proprietary \texttt{GPT-5-mini}~\cite{openai2025gpt5} model under the \emph{minimal} reasoning configuration (see OpenAI model documentation).

\subsection{Evaluation}

For each test trajectory, we evaluate model predictions at all possible starting positions \(t > 2\). Given a partial history \(u^i_{\le t}\), we autoregressively roll out the model for up to 5 steps using greedy-decoding. At each rollout step \(j \in \{1, \ldots, 5\}\), the model generates a submission \(\hat{a}^i_{t+j}\) conditioned on the original history and all previously generated rollout steps. Each generated submission is evaluated by the grading environment to produce the outcome \(\hat{o}^i_{t+j} = v^{(d^i)}_f(\hat{a}^i_{t+j})\), which is then appended to the context for subsequent predictions. Rollout terminates early if there is no corresponding ground truth student generation or if the model generates a fully correct solution.

\textbf{Metrics.} We report three complementary metrics averaged across all predictions.
First, \textbf{coverage} measures the proportion of ground-truth student submissions with model predictions. Coverage is less than 1.0 when models prematurely submit a correct solution (at step $k$ where $v^{(d^i)}_r(\hat{a}^i_{t+k}) = 1$) while the student has not yet finished ($v^{(d^i)}_r(a^i_{t+k}) < 1$), a known challenge in student simulation~\cite{miroyan2025parastudentgeneratingevaluatingrealistic}.
For matched pairs where models did generate, we evaluate generation quality using two metrics:
(1) \textbf{CodeBLEU}~\cite{ren2020codebleu}, measuring token-level syntactic and semantic similarity~\cite{duan2025automatedknowledgecomponentgeneration,duan2025testcase,liu-etal-2022-open}; and
(2) \textbf{grade proximity}, computed as \(1 - |v^{(d)}_r(\hat{a}^i_{t+j}) - v^{(d)}_r(a^i_{t+j})|\), where higher values indicate closer functional alignment. These quality metrics are computed only on matched pairs where models did generate, to avoid biasing results against models that correctly stop at perfect grades. Conversely, this means models with different coverage levels have their quality metrics computed on different susbet of predictions, which should be kept in mind when comparing across methods.
To assess robustness over rollouts, we also report \textbf{average degradation} $\Delta$ for each metric, which we compute as $\Delta m = \frac{1}{K-1}\sum_{k=2}^{K}(m_k - m_1)$, where $m_k$ denotes metric performance at rollout step $k$ and $K=5$. More negative values indicate faster degradation from the first step.

%% file: sections/05_results.tex
\section{Results}

\begin{figure}[t]
\centering
    \includegraphics[width=\linewidth,alt={Six line plots arranged in a two-by-three grid showing coverage, grade proximity, and CodeBLEU for Qwen3-4B (left column) and Qwen3-8B (right column) across rollout steps one through five. Fine-tuned methods including SFT, DPO, and DAPO maintain higher and more stable performance than the BASE and GPT-5-mini baselines, whose coverage drops sharply over rollout steps.}]{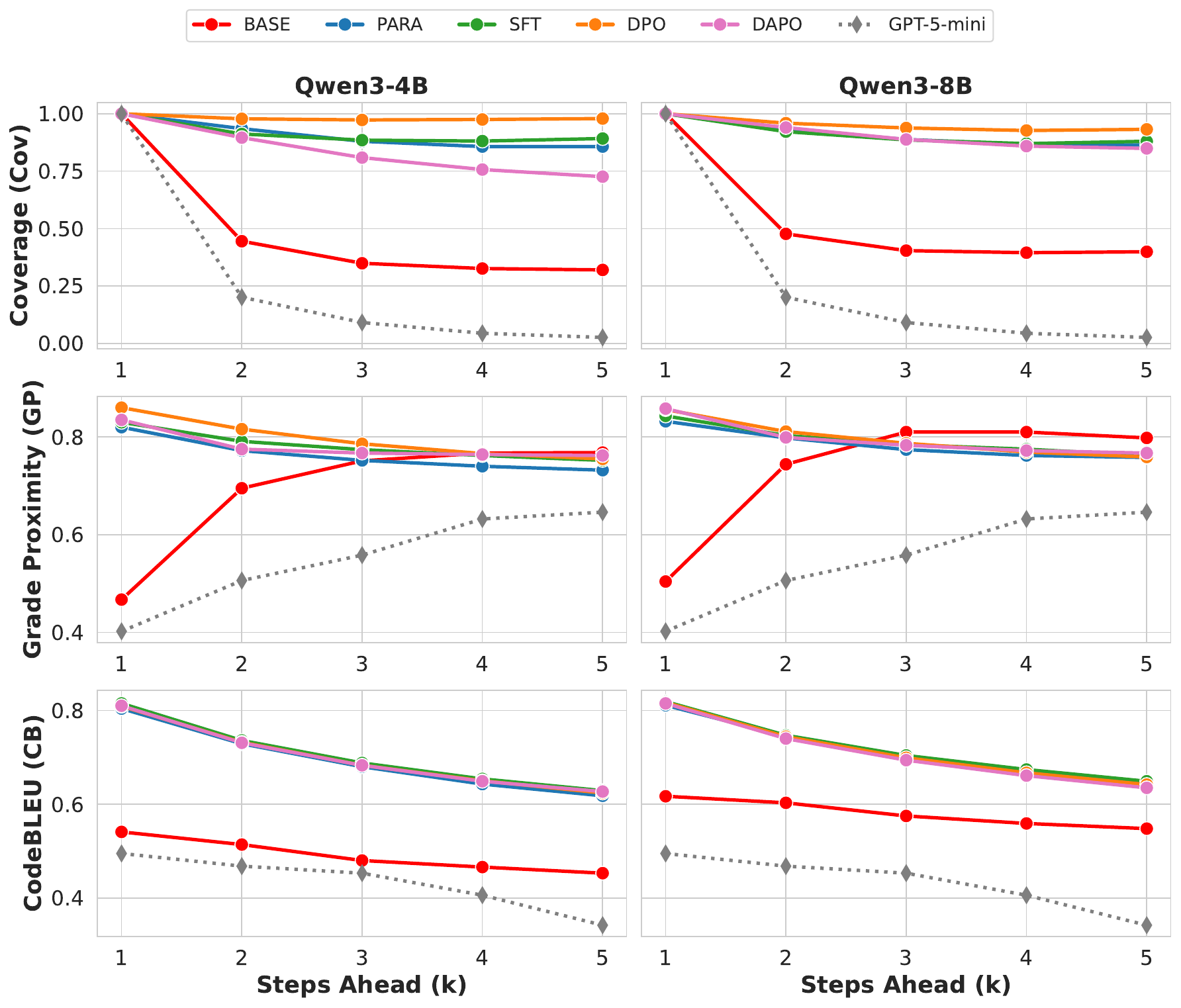}
    \caption{\textbf{Performance across rollout steps.} Coverage (top), Grade Proximity (middle) and CodeBLEU (bottom) as a function of autoregressive steps ahead (k=1 to k=5).}
    \label{fig:degradation}
\end{figure}

\begin{figure}[t]
    \centering
    \includegraphics[width=\linewidth,alt={Two line plots showing predicted grade as a function of normalized trajectory position for Qwen3-4B (left) and Qwen3-8B (right). Real student grades rise gradually from low values and jump sharply to one near the trajectory end. DPO and SFT models track this progression most closely, while BASE Qwen and GPT-5-mini predict near-constant grades independent of position.}]{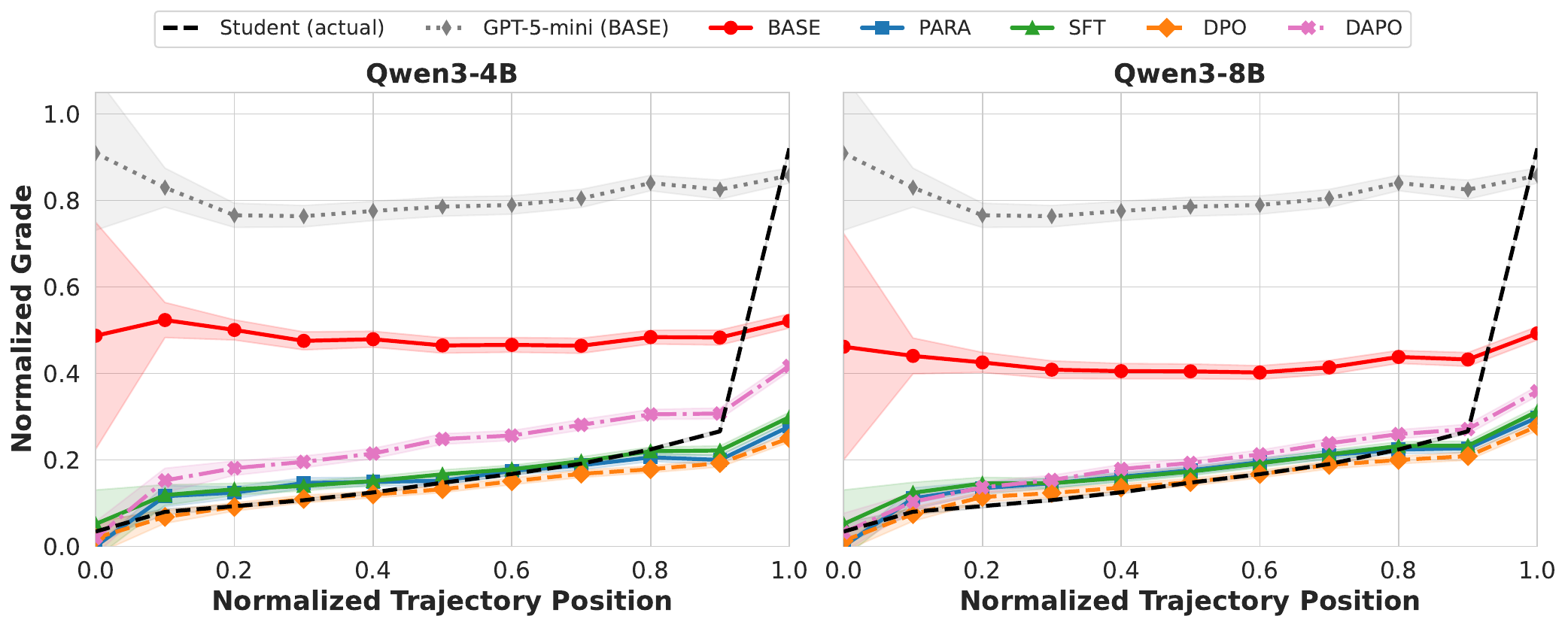}
    \caption{\textbf{Grade progression across normalized trajectory position}. Comparison of predicted grades between artificial student models and real students (black dashed, the average). The x-axis represents normalized position (by length) within a student's problem-solving trajectory (0 = start, 1 = end). Shaded regions indicate 95\% confidence intervals. The ground truth line does not reach 1.0 as some students failed. The abrupt jump reflects the proportion of students who resolve multiple failing test cases in a single revision.}
    \label{fig:progression}
\end{figure}

Table~\ref{tab:dataset_stats} summarizes our training and test split statistics. Table~\ref{tab:rollout_degradation} reports coverage and generation quality metrics averaged across all rollout steps. Figure~\ref{fig:degradation} details performance at each rollout step, and Figure~\ref{fig:progression} illustrates how model-generated grades evolve compared to students ground truth grades. We highlight several key findings below.

\input{table/statistics} 

\input{table/results}

\textbf{Prompted baselines cannot simulate realistic student behavior.}
BASE Qwen models achieve substantially lower coverage (0.528 for 4B, 0.570 for 8B), grade proximity (0.690, 0.733), and CodeBLEU (0.491, 0.581) than any fine-tuned model.
GPT-5-mini performs worst overall, with a coverage of only 0.340 and grade proximity of 0.549. Its low coverage indicates that it frequently generates fully correct solutions that terminate trajectories after one or two steps, while its low CodeBLEU (0.433) confirms that the code it does produce has little resemblance to student submissions. Interestingly, BASE Qwen models outperform GPT-5-mini, possibly because their smaller capacity leads to genuine mistakes that incidentally resemble student errors, whereas GPT-5-mini's stronger coding ability defaults to expert-like solutions despite the explicit personification prompt.

\textbf{Environment feedback improves student simulation.} SFT performs as well as or better than PARA on all three metrics for both models (Table~\ref{tab:rollout_degradation}). The margins are modest but the pattern is consistent. This is notable because PARA has a structural advantage: without environment feedback in the context, it can fit more prior student submissions within the same token budget. Despite this, environment feedback proves more informative than additional code history. We hypothesize that earlier submissions, particularly from already-passed test stages, carry diminishing signal about the student's current debugging strategy, whereas grader feedback directly conditions the next revision.

\textbf{Offline preference optimization yields the strongest results.} DPO achieves the highest coverage across both model sizes (0.982 for 4B, 0.957 for 8B), producing valid code at more rollout steps than any other method. It also achieves the highest grade proximity for 4B (0.797 vs.\ 0.782 for SFT). SFT achieves slightly higher CodeBLEU (0.704 vs.\ 0.700 for 4B; 0.718 vs.\ 0.714), although this gap may partly reflect DPO's higher coverage: by generating at more rollout steps, DPO is evaluated on a broader and likely harder subset of predictions. DAPO shows mixed results: it matches DPO on grade proximity for 8B but achieves the lowest coverage among fine-tuned models on 4B (0.859), and does not consistently improve over SFT. 


\textbf{Performance over steps and trajectory position.}
Looking at Figure~\ref{fig:degradation}, BASE model coverage drops sharply ($\Delta = -0.640$ for 4B, $-0.581$ for 8B), as they frequently generate solutions that terminate the trajectory prematurely. Trained models degrade less, with DPO maintaining the highest coverage throughout ($-0.024$ for 4B, $-0.061$ for 8B), while DAPO coverage reduces more substantially over time ($-0.203$ for 4B).
All fine-tuned models exhibit similar grade proximity degradation across rollout steps, with drops of approximately 0.06--0.08 from $k{=}1$ to $k{=}5$. The pattern over size is nuanced: for the 4B model, PARA degrades faster than SFT on grade proximity ($-0.071$ vs.\ $-0.060$), but for 8B, PARA degrades the least ($-0.059$) while preference-optimized models degrade the most (DPO: $-0.074$, DAPO: $-0.078$). Despite this, DPO maintains a higher absolute grade proximity compared to PARA at every rollout depth. Figure~\ref{fig:progression} shows that DPO-trained models most closely follow real student grade progression throughout normalized trajectories. DAPO models in contrast struggle to track students progressions. 
BASE models exhibit the opposite pattern: their grade proximity \textit{improves} over rollout steps ($+0.278$ for 4B, $+0.287$ for 8B), and GPT-5-mini shows the same trend ($+0.184$). Figure~\ref{fig:progression} reveals an interesting pattern: untrained models generate solutions that achieve similar grades independently of trajectory position. We hypothesize that these models attempt to generate correct solutions, rather than adapting to the student's current progress. The difference is that GPT-5-mini largely succeeds, producing near-perfect grades throughout, while BASE Qwen models fail enough test cases that the corresponding grades happen to align with students grades at later trajectory positions. Neither reproduces the progressive improvement characteristic of real students.
CodeBLEU shows more uniform downward trends across all trained models ($\approx -0.13$), suggesting that syntactic divergence accumulates at a similar rate regardless of training method. BASE model CodeBLEU remains comparatively stable ($-0.063$ for 4B, $-0.046$ for 8B).

%% file: table/statistics.tex
\begin{table}[h]
\centering
\caption{\textbf{FalconCode dataset statistics.} Summary of training and test splits. \#Traj: total trajectories; \#Stud: unique students; \#Succ/\#Fail: trajectories with final grade 100\%/below 100\%; \#Asg: unique programming assignments; Avg.\,Len: mean submissions per trajectory;}
\label{tab:dataset_stats}
\resizebox{0.80\columnwidth}{!}{%
\setlength{\tabcolsep}{3pt}
\renewcommand{\arraystretch}{1}
\begin{tabular}{lcccccccc}
\toprule
Split &
\#Traj &
\#Stud &
\#Succ &
\#Fail &
\#Asg &
Avg.\,Len \\
\midrule
Train &
1762 &
448 &
1712 &
50 &
17 &
22.4 \\
Test &
1000 &
384 &
981 &
19 &
15 &
20.1 \\
\bottomrule
\end{tabular}
}
\end{table}

%% file: table/results.tex
\begin{table}[htbp]
\centering
\caption{\textbf{Rollout performance metrics.} Legend: Cov (Coverage), GP (Grade Proximity), CB (CodeBLEU). Metrics averaged across k=1 to k=5 steps ahead, with degradation ($\Delta$) showing the average drop from k=1 to k=2..5. Negative $\Delta$ values indicate performance worsens over longer rollouts. \textbf{Bold}: best performance within each model.}
\label{tab:rollout_degradation}
\resizebox{\columnwidth}{!}{%
\begin{tabular}{ll|ccc|ccc}
\toprule
 &  & \multicolumn{3}{c|}{Average} & \multicolumn{3}{c}{Degradation} \\
Model & Method & Cov & GP & CB & $\Delta$Cov & $\Delta$GP & $\Delta$CB \\
\midrule
\multirow[t]{5}{*}{Qwen3-4B} & BASE & 0.528 & 0.690 & 0.491 & -0.640 & \textbf{0.278} & \textbf{-0.063} \\
 & PARA & 0.918 & 0.763 & 0.695 & -0.118 & -0.071 & -0.137 \\
 & SFT & 0.922 & 0.782 & \textbf{0.704} & -0.107 & -0.060 & -0.138 \\
 & DPO & \textbf{0.982} & \textbf{0.797} & 0.700 & \textbf{-0.024} & -0.079 & -0.137 \\
 & DAPO & 0.859 & 0.781 & 0.700 & -0.203 & -0.068 & -0.138 \\
\cline{1-8}
\multirow[t]{5}{*}{Qwen3-8B} & BASE & 0.570 & 0.733 & 0.581 & -0.581 & \textbf{0.287} & \textbf{-0.046} \\
 & PARA & 0.921 & 0.785 & 0.714 & -0.112 & -0.059 & -0.121 \\
 & SFT & 0.921 & 0.793 & \textbf{0.718} & -0.110 & -0.062 & -0.126 \\
 & DPO & \textbf{0.957} & \textbf{0.796} & 0.714 & \textbf{-0.061} & -0.074 & -0.129 \\
 & DAPO & 0.919 & \textbf{0.796} & 0.709 & -0.116 & -0.078 & -0.133 \\
\cline{1-8}
GPT-5-mini & BASE & \textbf{0.340} & \textbf{0.549} & \textbf{0.433} & \textbf{-0.910} & \textbf{0.184} & \textbf{-0.078} \\
\bottomrule
\end{tabular}
}
\end{table}

%% file: sections/06_concluding_discussion.tex
\section{Concluding Discussion}

Our experiments show that conversational serialization of student--environment interactions, combined with preference optimization, produces artificial students that more closely track real learners' debugging behavior than prompted baselines and models trained without feedback. Performance improvements compared to baselines are consistent in direction across both model sizes and all rollout steps. 
We believe coverage and grade proximity are particularly important dimensions for student simulation: models that generate code at every trajectory step, passing and failing the same tests at the same stages as real students, capture the incremental nature of real problem-solving more faithfully than models optimizing for surface-level code similarity alone. Among our findings, offline preference optimization via DPO provides the largest gains on these dimensions, achieving the highest coverage and grade proximity across both model sizes. Online preference optimization (DAPO) on the other hand does not consistently improve over SFT.
We hypothesize that DAPO's weaker results may stem from (a) our computational constraint of sampling only 2 prefixes per trajectory compared to DPO's exhaustive coverage, and (b) the absence of chain-of-thought reasoning~\cite{wei2022chainofthought} prior to code generation, which online methods typically rely on for effective optimization. Our results thus suggest that without explicit reasoning traces, offline methods may remain more practical than recent online alternatives for student simulation.

While our experiments focus on within-assignment simulation, we hope the proposed framework will support several existing research directions.

\textbf{Bridge with knowledge tracing and tutoring.}
Knowledge tracing (KT) models estimate a learner's latent proficiency across assignments, often predicting whether a student will solve a future task on the first attempt. Our framework operates at a complementary granularity: modeling the fine-grained debugging steps and intermediate attempts that precede success within a single task. We envision connecting these two levels by using cross-problem KC estimates as conditioning inputs to our simulation prompt~\cite{duan2026kaserknowledgealignedstudenterror}, allowing our artificial students to exhibit more realistic learner behavior.

\textbf{Bridge with tutoring.}
Moreover, our conversational serialization treats the automated assessment system as a dialogue partner that provides summative feedback. This is structurally identical to how a tutor agent would interact with a student. By representing student--environment interactions in this format, we could seamlessly extend sequences to include student--tutor exchanges, combining debugging traces with tutoring interactions~\cite{ashok-kumar-lan-2024-improving} in a single sequence. This opens several directions: our models could be adapted to predict student success conditioned jointly on debugging attempts and tutor exchanges~\cite{scarlatos2025training}, or integrated into conversational knowledge tracing frameworks~\cite{scarlatos2024exploringknowledgetracingtutorstudent}. More broadly, artificial students trained with our approach could serve as scalable simulators for reinforcement-learning--based tutor training~\cite{dinucu-jianu-etal-2025-problem}, where student success provides a training signal of tutoring strategy effectiveness.

\textbf{Limitations.}
Our work is not free of limitations. We experimented with a single Python programming dataset and one model family across two sizes. Validation on additional datasets from different institutional contexts (with different autograder feedback styles) and programming languages is needed. Our evaluation data is heavily skewed toward successful students: 95\% of trajectories end with a perfect score, meaning our results primarily reflect simulation quality for learners who ultimately succeed. This skew is partly inherent to programming courses, where most students eventually pass~\cite{watson2014failure} and those who do not often abandon the course entirely~\cite{bennedsen2019failure,koutcheme2022methodological}. However, struggling students are precisely those most in need of intervention; modeling their behavior, including trajectory abandonment, remains important future work. We also report results using greedy decoding on a single semester of test data without statistical testing; evaluating under varied sampling strategies and data splits would strengthen confidence in our findings. Finally, while our metrics demonstrate that models replicate student grade trajectories and code similarity, one needs to validate whether these new artificial students will be effectively useful for the downstream applications we motivate, such as tutor training or intervention testing. Establishing this educational validity is important future work.

\textbf{Future work.} 
Beyond addressing the highlighted limitations, there are three direct next steps for student simulation. First, we will explore agent-based artificial students equipped with execution, submission, and editing tools, mirroring the interactive environments real students use when programming. Second, we will improve our online preference optimization method. GRPO-style methods are known to improve model performance when elicited to produce chain-of-thought reasoning. By prompting models to generate explicit student-like thought traces in between tool operations, we hope to improve such models' ability to follow students' processes while potentially yielding post-analysis insights into why students code the way they do. Lastly, we will explore methods to address the challenge of efficiently integrating students' prior history (i.e., their attempts at prior assignments) to enable better personalization.

%% file: sections/07_acknowledgements.tex
\section*{Acknowledgments}

This work was supported by Research Council of Finland grants \#367787 and \#356114. We acknowledge the computational resources provided by the Aalto Science-IT project.

%% file: main.bib
@techreport{yang2025qwen3technicalreport,
	title        = {Qwen3 Technical Report},
	author       = {Qwen Team},
	year         = 2025,
	url          = {https://arxiv.org/abs/2505.09388},
	eprint       = {2505.09388},
	archiveprefix = {arXiv},
	primaryclass = {cs.CL}
}

@misc{ren2020codebleu,
	title        = {CodeBLEU: a Method for Automatic Evaluation of Code Synthesis},
	author       = {Shuo Ren and Daya Guo and Shuai Lu and Long Zhou and Shujie Liu and Duyu Tang and others},
	year         = 2020,
	eprint       = {2009.10297},
	archiveprefix = {arXiv},
	primaryclass = {cs.SE}
}

@article{ouyang2022training,
	title        = {Training language models to follow instructions with human feedback},
	author       = {Ouyang, Long and Wu, Jeffrey and Jiang, Xu and Almeida, Diogo and Wainwright, Carroll and Mishkin, Pamela and others},
	year         = 2022,
	journal      = {Advances in neural information processing systems},
	volume       = 35,
	pages        = {27730--27744}
}

@misc{le_coderl_2022,
	title        = {{CodeRL}: {Mastering} {Code} {Generation} through {Pretrained} {Models} and {Deep} {Reinforcement} {Learning}},
	shorttitle   = {{CodeRL}},
	author       = {Le, Hung and Wang, Yue and Gotmare, Akhilesh Deepak and Savarese, Silvio and Hoi, Steven C. H.},
	year         = 2022,
	month        = nov,
	booktitle    = {Advances in Neural Information Processing Systems},
	publisher    = {arXiv},
	volume       = 35,
	pages        = {21314--21328},
	url          = {http://arxiv.org/abs/2207.01780},
	urldate      = {2023-01-04},
	note         = {arXiv:2207.01780 [cs]}
}

@inproceedings{dettmers2023qloraefficientfinetuningquantized,
	title        = {QLORA: Efficient Finetuning of Quantized LLMs},
	author       = {Tim Dettmers and Artidoro Pagnoni and Ari Holtzman and Luke Zettlemoyer},
	year         = 2023,
	booktitle    = {Proceedings of the 37th International Conference on Neural Information Processing Systems},
	location     = {New Orleans, LA, USA},
	publisher    = {Curran Associates Inc.},
	address      = {Red Hook, NY, USA},
	series       = {NIPS '23},
	pages        = 441,
	numpages     = 28
}

@article{Guo2025,
	title        = {DeepSeek-R1 incentivizes reasoning in LLMs through reinforcement learning},
	author       = {Guo, Daya and Yang, Dejian and Zhang, Haowei and Song, Junxiao and Wang, Peiyi and Zhu, Qihao and others},
	year         = 2025,
	journal      = {Nature},
	volume       = 645,
	pages        = {633--638},
	doi          = {10.1038/s41586-025-09422-z},
	url          = {https://doi.org/10.1038/s41586-025-09422-z}
}

@misc{yu2025dapoopensourcellmreinforcement,
	title        = {DAPO: An Open-Source LLM Reinforcement Learning System at Scale},
	author       = {Qiying Yu and Zheng Zhang and Ruofei Zhu and Yufeng Yuan and Xiaochen Zuo and Yu Yue and others},
	year         = 2025,
	url          = {https://arxiv.org/abs/2503.14476},
	eprint       = {2503.14476},
	archiveprefix = {arXiv},
	primaryclass = {cs.LG}
}

@inproceedings{opencodeinterpreter,
	title        = {OpenCodeInterpreter: Integrating Code Generation with Execution and Refinement},
	author       = {Tianyu Zheng and Ge Zhang and Tianhao Shen and Xueling Liu and Bill Yuchen Lin and Jie Fu and others},
	year         = 2024,
	booktitle    = {Findings of the Association for Computational Linguistics, {ACL} 2024, Bangkok, Thailand and virtual meeting, August 11-16, 2024},
	publisher    = {Association for Computational Linguistics},
	pages        = {12834--12859},
	doi          = {10.18653/V1/2024.FINDINGS-ACL.762},
	url          = {https://doi.org/10.18653/v1/2024.findings-acl.762},
	bibsource    = {dblp computer science bibliography, https://dblp.org}
}

@article{schulman2025lora,
	title        = {LoRA Without Regret},
	author       = {John Schulman and Thinking Machines Lab},
	year         = 2025,
	journal      = {Thinking Machines Lab: Connectionism},
	doi          = {10.64434/tml.20250929},
	note         = {https://thinkingmachines.ai/blog/lora/}
}

@misc{unsloth,
	title        = {Unsloth},
	author       = {Daniel Han, Michael Han and Unsloth team},
	year         = 2023,
	url          = {http://github.com/unslothai/unsloth}
}

@misc{unsloth_lora_hyperparameters,
	title        = {LoRA Hyperparameters Guide},
	author       = {{Unsloth AI}},
	year         = 2024,
	url          = {https://docs.unsloth.ai/get-started/fine-tuning-llms-guide/lora-hyperparameters-guide},
	note         = {Accessed: 2025-12-23},
	organization = {Unsloth}
}

@inproceedings{rafailov2024direct,
	title        = {Direct preference optimization: your language model is secretly a reward model},
	author       = {Rafailov, Rafael and Sharma, Archit and Mitchell, Eric and Ermon, Stefano and Manning, Christopher D. and Finn, Chelsea},
	year         = 2023,
	booktitle    = {Proceedings of the 37th International Conference on Neural Information Processing Systems},
	location     = {New Orleans, LA, USA},
	publisher    = {Curran Associates Inc.},
	address      = {Red Hook, NY, USA},
	series       = {NIPS '23},
	articleno    = 2338,
	numpages     = 14
}

@techreport{openai2025gpt5,
	title        = {{GPT-5 System Card}},
	author       = {{OpenAI}},
	year         = 2025,
	url          = {https://openai.com/index/gpt-5-system-card},
	institution  = {OpenAI}
}

@inproceedings{wei2022chainofthought,
	title        = {Chain-of-Thought Prompting Elicits Reasoning in Large Language Models},
	author       = {Wei, Jason and Wang, Xuezhi and Schuurmans, Dale and Bosma, Maarten and Ichter, Brian and Xia, Fei and others},
	year         = 2022,
	booktitle    = {Advances in Neural Information Processing Systems},
	volume       = 35,
	pages        = {24824--24837}
}

@article{mukherjee2023orca,
	title        = {Orca: Progressive Learning from Complex Explanation Traces of {GPT-4}},
	author       = {Mukherjee, Subhabrata and Mitra, Arindam and Jawahar, Ganesh and Agarwal, Sahaj and Palangi, Hamid and Awadallah, Ahmed},
	year         = 2023,
	journal      = {arXiv preprint arXiv:2306.02707}
}

@inproceedings{ding2023enhancing,
	title        = {Enhancing Chat Language Models by Scaling High-quality Instructional Conversations},
	author       = {Ding, Ning and Chen, Yulin and Xu, Bokai and Qin, Yujia and Hu, Shengding and Liu, Zhiyuan and others},
	year         = 2023,
	booktitle    = {Proceedings of the 2023 Conference on Empirical Methods in Natural Language Processing},
	pages        = {3029--3051}
}

@inproceedings{xu2024wizardlm,
	title        = {{WizardLM}: Empowering Large Pre-Trained Language Models to Follow Complex Instructions},
	author       = {Xu, Can and Sun, Qingfeng and Zheng, Kai and Geng, Xiubo and Zhao, Pu and Feng, Jiazhan and others},
	year         = 2024,
	booktitle    = {The Twelfth International Conference on Learning Representations}
}

@inproceedings{akter2025mind,
	title        = {{MIND}: Math Informed syNthetic Dialogues for Pretraining {LLM}s},
	author       = {Akter, Syeda Nahida and Prabhumoye, Shrimai and Kamalu, John and Satheesh, Sanjeev and Nyberg, Eric and Patwary, Mostofa and others},
	year         = 2025,
	booktitle    = {The Thirteenth International Conference on Learning Representations}
}

@inproceedings{react,
	title        = {ReAct: Synergizing Reasoning and Acting in Language Models},
	author       = {Shunyu Yao and Jeffrey Zhao and Dian Yu and Nan Du and Izhak Shafran and Karthik R. Narasimhan and others},
	year         = 2023,
	booktitle    = {The Eleventh International Conference on Learning Representations, {ICLR} 2023, Kigali, Rwanda, May 1-5, 2023},
	publisher    = {OpenReview.net},
	url          = {https://openreview.net/forum?id=WE\%5FvluYUL-X},
	bibsource    = {dblp computer science bibliography, https://dblp.org}
}

@inproceedings{falconcode,
	title        = {FalconCode: A Multiyear Dataset of Python Code Samples from an Introductory Computer Science Course},
	author       = {de Freitas, Adrian and Coffman, Joel and de Freitas, Michelle and Wilson, Justin and Weingart, Troy},
	year         = 2023,
	booktitle    = {Proceedings of the 54th ACM Technical Symposium on Computer Science Education V. 1},
	location     = {Toronto ON, Canada},
	publisher    = {Association for Computing Machinery},
	address      = {New York, NY, USA},
	series       = {SIGCSE 2023},
	pages        = {938–944},
	doi          = {10.1145/3545945.3569822},
	isbn         = 9781450394314,
	numpages     = 7
}

@inproceedings{macneil2024synthetic,
	title        = {Synthetic Students: A Comparative Study of Bug Distribution Between Large Language Models and Computing Students},
	author       = {MacNeil, Stephen and Rogalska, Magdalena and Leinonen, Juho and Denny, Paul and Hellas, Arto and Crosland, Xandria},
	year         = 2024,
	booktitle    = {Proceedings of the 2024 on ACM Virtual Global Computing Education Conference V. 1},
	location     = {Virtual Event, NC, USA},
	publisher    = {Association for Computing Machinery},
	address      = {New York, NY, USA},
	series       = {SIGCSE Virtual 2024},
	pages        = {137–143},
	doi          = {10.1145/3649165.3690100},
	isbn         = 9798400705984,
	url          = {https://doi.org/10.1145/3649165.3690100},
	numpages     = 7
}

@inproceedings{leinonen2025llmitation,
	title        = {LLM-itation is the Sincerest Form of Data: Generating Synthetic Buggy Code Submissions for Computing Education},
	author       = {Leinonen, Juho and Denny, Paul and Kiljunen, Olli and MacNeil, Stephen and Sarsa, Sami and Hellas, Arto},
	year         = 2025,
	booktitle    = {Proceedings of the 27th Australasian Computing Education Conference},
	location     = {},
	publisher    = {Association for Computing Machinery},
	address      = {New York, NY, USA},
	series       = {ACE '25},
	pages        = {56–63},
	doi          = {10.1145/3716640.3716647},
	isbn         = 9798400714252,
	url          = {https://doi.org/10.1145/3716640.3716647},
	numpages     = 8
}

@article{corbett1994knowledge,
	title        = {Knowledge tracing: Modeling the acquisition of procedural knowledge},
	author       = {Corbett, Albert T. and Anderson, John R.},
	year         = 1994,
	journal      = {User Modeling and User-Adapted Interaction},
	volume       = 4,
	number       = 4,
	pages        = {253--278}
}

@inproceedings{jadud2006methods,
	title        = {Methods and tools for exploring novice compilation behaviour},
	author       = {Jadud, Matthew C},
	year         = 2006,
	booktitle    = {Proceedings of the second international workshop on Computing education research},
	pages        = {73--84}
}

@inproceedings{watson2013predicting,
	title        = {Predicting performance in an introductory programming course by logging and analyzing student programming behavior},
	author       = {Watson, Christopher and Li, Frederick WB and Godwin, Jamie L},
	year         = 2013,
	booktitle    = {2013 IEEE 13th international conference on advanced learning technologies},
	pages        = {319--323},
	organization = {IEEE}
}

@inproceedings{sirkia2012exploring,
	title        = {Exploring programming misconceptions: an analysis of student mistakes in visual program simulation exercises},
	author       = {Sirki{\"a}, Teemu and Sorva, Juha},
	year         = 2012,
	booktitle    = {Proceedings of the 12th Koli calling international conference on computing education research},
	pages        = {19--28}
}

@inproceedings{azcona2017targeting,
	title        = {Targeting at-risk students using engagement and effort predictors in an introductory computer programming course},
	author       = {Azcona, David and Smeaton, Alan F},
	year         = 2017,
	booktitle    = {European Conference on Technology Enhanced Learning},
	pages        = {361--366},
	organization = {Springer}
}

@article{Butler1995FeedbackAS,
	title        = {Feedback and Self-Regulated Learning: A Theoretical Synthesis},
	author       = {Butler, Deborah L. and Winne, Philip H.},
	year         = 1995,
	journal      = {Review of Educational Research},
	volume       = 65,
	pages        = {245--281}
}

@inproceedings{watson2014failure,
author = {Watson, Christopher and Li, Frederick W.B.},
title = {Failure rates in introductory programming revisited},
year = {2014},
isbn = {9781450328333},
publisher = {ACM},
address = {New York, NY, USA},
url = {https://doi.org/10.1145/2591708.2591749},
doi = {10.1145/2591708.2591749},
booktitle = {Proceedings of the 2014 Conference on Innovation \& Technology in Computer Science Education},
pages = {39–44},
numpages = {6},
location = {Uppsala, Sweden},
series = {ITiCSE '14}
}

@article{bennedsen2019failure,
author = {Bennedsen, Jens and Caspersen, Michael E.},
title = {Failure rates in introductory programming: 12 years later},
year = {2019},
issue_date = {June 2019},
publisher = {Association for Computing Machinery},
address = {New York, NY, USA},
volume = {10},
number = {2},
issn = {2153-2184},
url = {https://doi.org/10.1145/3324888},
doi = {10.1145/3324888},
journal = {ACM Inroads},
month = apr,
pages = {30–36},
numpages = {7}
}

@inproceedings{koutcheme2022methodological,
author = {Koutcheme, Charles and Sarsa, Sami and Hellas, Arto and Haaranen, Lassi and Leinonen, Juho},
title = {Methodological Considerations for Predicting At-risk Students},
year = {2022},
isbn = {9781450396431},
publisher = {Association for Computing Machinery},
address = {New York, NY, USA},
url = {https://doi.org/10.1145/3511861.3511873},
doi = {10.1145/3511861.3511873},
booktitle = {Proceedings of the 24th Australasian Computing Education Conference},
pages = {105–113},
numpages = {9},
location = {Virtual Event, Australia},
series = {ACE '22}
}

@misc{miroyan2025parastudentgeneratingevaluatingrealistic,
	title        = {ParaStudent: Generating and Evaluating Realistic Student Code by Teaching LLMs to Struggle},
	author       = {Mihran Miroyan and Rose Niousha and Joseph E. Gonzalez and Gireeja Ranade and Narges Norouzi},
	year         = 2025,
	url          = {https://arxiv.org/abs/2507.12674},
	eprint       = {2507.12674},
	archiveprefix = {arXiv},
	primaryclass = {cs.CY}
}

@inproceedings{liu-etal-2022-open,
	title        = {Open-ended Knowledge Tracing for Computer Science Education},
	author       = {Liu, Naiming  and Wang, Zichao  and Baraniuk, Richard  and Lan, Andrew},
	year         = 2022,
	month        = dec,
	booktitle    = {Proceedings of the 2022 Conference on Empirical Methods in Natural Language Processing},
	publisher    = {Association for Computational Linguistics},
	address      = {Abu Dhabi, United Arab Emirates},
	pages        = {3849--3862},
	doi          = {10.18653/v1/2022.emnlp-main.254},
}

@inproceedings{duan2025testcase,
	title        = {Test Case-Informed Knowledge Tracing for Open-ended Coding Tasks},
	author       = {Duan, Zhangqi and Fernandez, Nigel and Hicks, Alexander and Lan, Andrew},
	year         = 2025,
	booktitle    = {Proceedings of the 15th International Learning Analytics and Knowledge Conference},
	location     = {},
	publisher    = {Association for Computing Machinery},
	address      = {New York, NY, USA},
	series       = {LAK '25},
	pages        = {238–248},
	doi          = {10.1145/3706468.3706500},
	isbn         = 9798400707018,
	url          = {https://doi.org/10.1145/3706468.3706500},
	numpages     = 11
}

@misc{duan2025automatedknowledgecomponentgeneration,
	title        = {Automated Knowledge Component Generation for Interpretable Knowledge Tracing in Coding Problems},
	author       = {Zhangqi Duan and Nigel Fernandez and Arun Balajiee Lekshmi Narayanan and Mohammad Hassany and Rafaella Sampaio de Alencar and Peter Brusilovsky and Bita Akram and others},
	year         = 2025,
	url          = {https://arxiv.org/abs/2502.18632},
	eprint       = {2502.18632},
	archiveprefix = {arXiv},
	primaryclass = {cs.AI}
}

@inbook{scarlatos2024improvingvalidityautomaticallygenerated,
	title        = {Improving the Validity of Automatically Generated Feedback via Reinforcement Learning},
	author       = {Scarlatos, Alexander and Smith, Digory and Woodhead, Simon and Lan, Andrew},
	year         = 2024,
	booktitle    = {Artificial Intelligence in Education},
	publisher    = {Springer Nature Switzerland},
	pages        = {280–294},
	doi          = {10.1007/978-3-031-64302-6_20},
	isbn         = 9783031643026,
	issn         = {1611-3349},
	organization = {Springer}
}

@inproceedings{koutcheme-etal-2025-direct,
	title        = {Direct Repair Optimization: Training Small Language Models For Educational Program Repair Improves Feedback},
	author       = {Koutcheme, Charles  and Dainese, Nicola  and Hellas, Arto},
	year         = 2025,
	month        = jul,
	booktitle    = {Proceedings of the 20th Workshop on Innovative Use of NLP for Building Educational Applications (BEA 2025)},
	publisher    = {Association for Computational Linguistics},
	address      = {Vienna, Austria},
	pages        = {564--581},
	doi          = {10.18653/v1/2025.bea-1.41},
	isbn         = {979-8-89176-270-1},
	url          = {https://aclanthology.org/2025.bea-1.41/}
}

@inproceedings{phung2024automatinghumantutorstyleprogramming,
	title        = {Automating Human Tutor-Style Programming Feedback: Leveraging GPT-4 Tutor Model for Hint Generation and GPT-3.5 Student Model for Hint Validation},
	author       = {Phung, Tung and P\u{a}durean, Victor-Alexandru and Singh, Anjali and Brooks, Christopher and Cambronero, Jos\'{e} and Gulwani, Sumit and Singla, Adish and others},
	year         = 2024,
	booktitle    = {Proceedings of the 14th Learning Analytics and Knowledge Conference},
	location     = {Kyoto, Japan},
	publisher    = {ACM},
	address      = {New York, NY, USA},
	series       = {LAK '24},
	pages        = {12–23},
	doi          = {10.1145/3636555.3636846},
	isbn         = 9798400716188,
	numpages     = 12
}

@inproceedings{woodrow2025dpo,
	title        = {Improving Generative AI Student Feedback: Direct Preference Optimization with Teachers in the Loop},
	author       = {Juliette Woodrow and Chris Piech and Sanmi Koyejo},
	year         = 2025,
	month        = {July},
	booktitle    = {Proceedings of the 18th International Conference on Educational Data Mining},
	publisher    = {International Educational Data Mining Society},
	address      = {Palermo, Italy},
	pages        = {442--449},
	doi          = {10.5281/zenodo.15870266},
	isbn         = {978-1-7336736-6-2},
	url          = {https://doi.org/10.5281/zenodo.15870266},
	venue        = {Palermo, Italy}
}

@inproceedings{wang2017deep,
	title        = {Deep knowledge tracing on programming exercises},
	author       = {Wang, Lisa and Sy, Angela and Liu, Larry and Piech, Chris},
	year         = 2017,
	booktitle    = {Proceedings of the fourth (2017) ACM conference on learning@ scale},
	pages        = {201--204}
}

@article{kasurinen2009estimating,
	title        = {Estimating programming knowledge with Bayesian knowledge tracing},
	author       = {Kasurinen, Jussi and Nikula, Uolevi},
	year         = 2009,
	journal      = {ACM SIGCSE Bulletin},
	publisher    = {ACM New York, NY, USA},
	volume       = 41,
	number       = 3,
	pages        = {313--317}
}

@article{brown1999simulated,
	title        = {Simulated classrooms and artificial students: The potential effects of new technologies on teacher education},
	author       = {Brown, Abbie Howard},
	year         = 1999,
	journal      = {Journal of research on computing in education},
	publisher    = {Taylor \& Francis},
	volume       = 32,
	number       = 2,
	pages        = {307--318}
}

@inproceedings{matsuda2007evaluating,
	title        = {Evaluating a simulated student using real students data for training and testing},
	author       = {Matsuda, Noboru and Cohen, William W and Sewall, Jonathan and Lacerda, Gustavo and Koedinger, Kenneth R},
	year         = 2007,
	booktitle    = {International Conference on User Modeling},
	pages        = {107--116},
	organization = {Springer}
}

@article{vanlehn1994applications,
  title     = {Applications of Simulated Students: An Exploration},
  author    = {VanLehn, Kurt and Ohlsson, Stellan and Nason, Rod},
  year      = 1994,
  journal   = {Journal of Artificial Intelligence in Education},
  volume    = 5,
  number    = 2,
  pages     = {135--175}
}

@inproceedings{dinucu-jianu-etal-2025-problem,
	title        = {From Problem-Solving to Teaching Problem-Solving: Aligning {LLM}s with Pedagogy using Reinforcement Learning},
	author       = {Dinucu-Jianu, David  and Macina, Jakub  and Daheim, Nico  and Hakimi, Ido  and Gurevych, Iryna  and Sachan, Mrinmaya},
	year         = 2025,
	month        = nov,
	booktitle    = {Proceedings of the 2025 Conference on Empirical Methods in Natural Language Processing},
	publisher    = {Association for Computational Linguistics},
	address      = {Suzhou, China},
	pages        = {272--292},
	doi          = {10.18653/v1/2025.emnlp-main.15},
	isbn         = {979-8-89176-332-6},
	url          = {https://aclanthology.org/2025.emnlp-main.15/}
}

@inproceedings{scarlatos2024exploringknowledgetracingtutorstudent,
	title        = {Exploring Knowledge Tracing in Tutor-Student Dialogues using LLMs},
	author       = {Alexander Scarlatos and Ryan S. Baker and Andrew Lan},
	year         = 2025,
	booktitle    = {Proceedings of the 15th Learning Analytics and Knowledge Conference, {LAK} 2025, Dublin, Ireland, March 3-7, 2025},
	publisher    = {{ACM}}
}

@inproceedings{scarlatos2025training,
	title        = {Training LLM-Based Tutors to Improve Student Learning Outcomes in Dialogues},
	author       = {Scarlatos, Alexander and Liu, Naiming and Lee, Jaewook and Baraniuk, Richard and Lan, Andrew},
	year         = 2025,
	booktitle    = {Artificial Intelligence in Education},
    editor="Cristea, Alexandra I.
    and Walker, Erin
    and Lu, Yu
    and Santos, Olga C.
    and Isotani, Seiji",
	publisher    = {Springer Nature Switzerland},
	address      = {Cham},
	pages        = {251--266},
	isbn         = {978-3-031-98414-3}
}

@inproceedings{ashok-kumar-lan-2024-improving,
	title        = {Improving Socratic Question Generation using Data Augmentation and Preference Optimization},
	author       = {Ashok Kumar, Nischal  and Lan, Andrew},
	year         = 2024,
	month        = jun,
	booktitle    = {Proceedings of the 19th Workshop on Innovative Use of NLP for Building Educational Applications (BEA 2024)},
	publisher    = {Association for Computational Linguistics},
	address      = {Mexico City, Mexico},
	pages        = {108--118},
	url          = {https://aclanthology.org/2024.bea-1.10/}
}

@misc{ross2025modelingstudentlearning38,
	title        = {Modeling Student Learning with 3.8 Million Program Traces},
	author       = {Alexis Ross and Megha Srivastava and Jeremiah Blanchard and Jacob Andreas},
	year         = 2025,
	url          = {https://arxiv.org/abs/2510.05056},
	eprint       = {2510.05056},
	archiveprefix = {arXiv},
	primaryclass = {cs.LG}
}

@inproceedings{PyTaskSyn,
	title        = {Synthesizing High-Quality Programming Tasks with LLM-Based Expert and Student Agents},
	author       = {Nguyen, Manh Hung and P{\u{a}}durean, Victor-Alexandru and Gotovos, Alkis and Tschiatschek, Sebastian and Singla, Adish},
	year         = 2025,
	booktitle    = {Artificial Intelligence in Education},
	publisher    = {Springer Nature Switzerland},
	address      = {Cham},
	pages        = {77--91},
	isbn         = {978-3-031-98414-3},
    editor="Cristea, Alexandra I.
    and Walker, Erin
    and Lu, Yu
    and Santos, Olga C.
    and Isotani, Seiji"
}

@misc{duan2026kaserknowledgealignedstudenterror,
	title        = {KASER: Knowledge-Aligned Student Error Simulator for Open-Ended Coding Tasks},
	author       = {Zhangqi Duan and Nigel Fernandez and Andrew Lan},
	year         = 2026,
	url          = {https://arxiv.org/abs/2601.06633},
	eprint       = {2601.06633},
	archiveprefix = {arXiv},
	primaryclass = {cs.LG}
}

@inproceedings{koutcheme2026aligning,
author = {Koutcheme, Charles and Woodrow, Juliette and Piech, Chris},
title = {Aligning Small Language Models for Programming Feedback: Towards Scalable Coding Support in a Massive Global Course},
year = {2026},
isbn = {9798400722561},
publisher = {Association for Computing Machinery},
address = {New York, NY, USA},
url = {https://doi.org/10.1145/3770762.3772539},
doi = {10.1145/3770762.3772539},
booktitle = {Proceedings of the 57th ACM Technical Symposium on Computer Science Education V.1},
pages = {610–616},
numpages = {7},
location = {USA},
series = {SIGCSE TS 2026}
}
